\documentclass[11pt,a4paper]{article}
\usepackage[hyperref]{acl2021}
\usepackage{times}
\usepackage{latexsym}
\usepackage{xcolor}
\usepackage{graphicx}
\usepackage{amsmath, bm}
\usepackage{amsfonts}
\usepackage{multirow}
\usepackage{cleveref}
\usepackage{booktabs}
\usepackage{xcolor}
\usepackage{enumitem}
\usepackage{amssymb}
\usepackage{pifont}

\DeclareMathOperator*{\argmax}{arg\,max}

\newcommand{\seq}[1]{\boldsymbol{#1}}

\usepackage{algorithm}
\usepackage{algorithmicx}
\usepackage{algpseudocode}

\algnewcommand\algorithmicinput{\textbf{Input:}}
\algnewcommand\INPUT{\item[\algorithmicinput]}
\algnewcommand\algorithmicoutput{\textbf{Output:}}
\algnewcommand\OUTPUT{\item[\algorithmicoutput]}

\usepackage{microtype}

\aclfinalcopy 

\title{GWLAN: General Word-Level AutocompletioN \\ for Computer-Aided Translation}

\author{Huayang Li~~~Lemao Liu~~~Guoping Huang~~~Shuming Shi \\
        Tencent AI Lab\\
        \tt \{alanili,redmondliu,donkeyhuang,shumingshi\}@tencent.com 
}
\date{}

\begin{document}
\maketitle
\begin{abstract}

Computer-aided translation (CAT), the use of software to assist a human translator in the translation process, has been proven to be useful in enhancing the productivity of human translators. Autocompletion, which suggests translation results according to the text pieces provided by human translators, is a core function of CAT. There are two limitations in previous research in this line. First, most research works on this topic focus on sentence-level autocompletion (i.e., generating the whole translation as a sentence based on human input), but word-level autocompletion is under-explored so far. Second, almost no public benchmarks are available for the autocompletion task of CAT. This might be among the reasons why research progress in CAT is much slower compared to automatic MT. In this paper, we propose the task of general word-level autocompletion (GWLAN) from a real-world CAT scenario, and construct the first public benchmark\footnote{The information of benchmark datasets is in \url{https://github.com/ghrua/gwlan}} to facilitate research in this topic. In addition, we propose an effective method for GWLAN and compare it with several strong baselines. Experiments demonstrate that our proposed method can give significantly more accurate predictions than the baseline methods on our benchmark datasets.

\end{abstract}

\section{Introduction}

\begin{figure}
\begin{center}
   \includegraphics[width=1.0\linewidth]{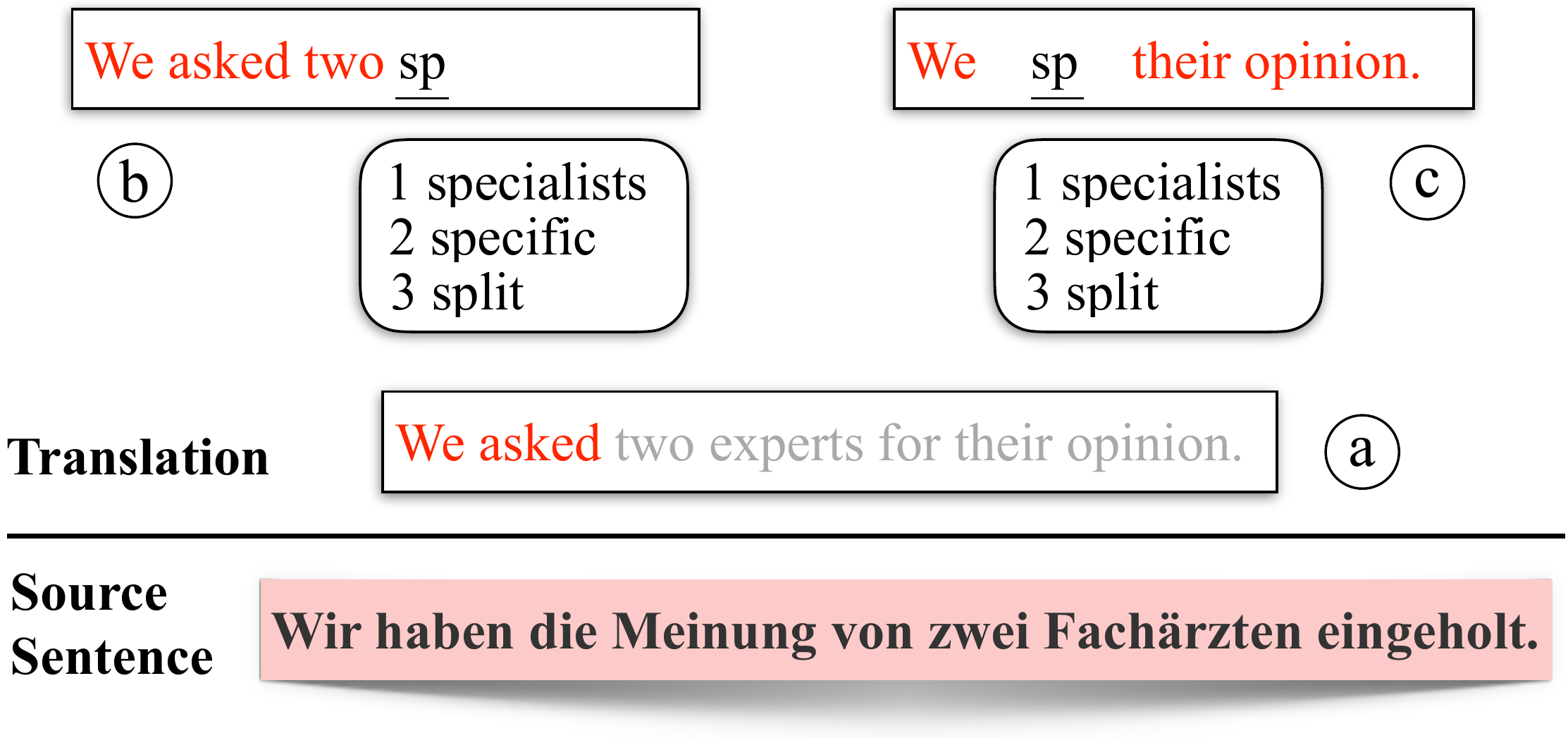}
\end{center}
   \caption{Illustration of Different Autocompletion Methods. The translation context is in \textcolor{red}{red}. Sub-figure in (a) is the sentence-level autocompletion, where the \textcolor{gray}{gray part} is the completion generated by MT system. Both (b) and (c) are word-level autocompletion, underlined text ``sp'' is the human typed characters and the words in the rounded rectangles are word-level autocompletion candidates.}
\label{fig:illustration}
\end{figure}

Machine translation (MT) has witnessed great advancements with the emergence of neural machine translation (NMT) \cite{sutskever2014sequence,bahdanau2014neural,wu2016google,gehring2017convolutional,vaswani2017attention}, which is able to produce much higher quality translation results than statistical machine translation (SMT) models~\cite{koehn-etal-2003-statistical,chiang2005hierarchical,koehn2009statistical}. In spite of this, MT systems cannot replace human translators, especially in the scenarios with rigorous translation quality requirements (e.g., translating product manuals, patent documents, government policies, and other official documents).
Therefore, how to leverage the pros of MT systems to help human translators, namely, \textit{Computer-aided translation} (CAT), attracts the attention of  researchers \cite{barrachina2009statistical, green2014human, knowles2016neural,santy-etal-2019-inmt}. Among all CAT technologies (such as translation memory, terminology management, sample sentence search, etc.), \textit{autocompletion} plays an important role in a CAT system in enhancing translation efficiency. Autocompletion suggests translation results according to the text pieces provided by human translators.

We note two limitations in previous research on the topic of autocompletion for CAT.
First, most of previous studies aim to save human efforts by sentence-level autocompletion (Figure \ref{fig:illustration} a). Nevertheless, word-level autocompletion (Figure \ref{fig:illustration} b and c) has not been systematically studied.
Second, almost no public benchmarks are available for the autocompletion task of CAT.
Although some achievements have been made, research progress in CAT is more sluggish than that in automatic MT. The lack of benchmarks has hindered researchers from making continuous progress in this area.

In this work, we propose a \textbf{G}eneral {\bf W}ord-{\bf L}evel {\bf A}utocompletio{\bf N} (GWLAN) task, and construct a benchmark with automatic evaluation to facilitate further research progress in CAT.
Specifically, the GWLAN task aims to complete the target word for human translators based on a source sentence, translation context as well as human typed characters. Compared with previous work, GWLAN considers four most general types of translation context: prefix, suffix, zero context, and bidirectional context. Besides, as in most real world scenarios, we only know the relative position between input words and the spans of translation context in the GWLAN task.
We construct a benchmark for the task, with the goal of supporting automatic evaluation and ensuring a convenient and fair comparison among different methods. The benchmark is built by extracting triples of source sentences, translation contexts, and human typed characters from standard parallel datasets. Accuracy is adopted as the evaluation metric in the benchmark.

To address the variety of context types and weak position information issue, we propose a neural model to complete a word in different types of context as well as a joint training strategy to optimize its parameters. Our model can learn the representation of potential target words in translation and then choose the most possible word based on the human input.

Our contributions are two-fold:

\begin{itemize}
    \item We propose the task of general word-level autocompletion for CAT, and construct the first public benchmark to facilitate research in this topic.
    \item We propose a joint training strategy to optimize the model parameters on different types of contexts together.~\footnote{This approach has been implemented into a human-machine interactive translation system TranSmart~\cite{transmart2021} at \url{www.transmart.qq.com}.} 
\end{itemize}



\section{Related Work}

Computer-aided translation (CAT) is a widely used practice when using MT technology in the industry. As the the MT systems advanced and improved, various efficient interaction ways of CAT have emerged \cite{vasconcellos1985spanam, green2014human, hokamp2017lexically,weng2019correct,wang+:2020:touch}.  Among those different methods, the autocompletion is the most related to our work. Therefore, we will first describe previous works in both sentence-level and word-level autocompletion, then show the relation to other tasks and scenarios.

\paragraph{Sentence-level Autocompletion} Most of previous work in autocompletion for CAT focus on sentence-level completion. A common use case in this line is interactive machine translation (IMT) \cite{green2014human,cheng2016primt,peris2017interactive, knowles2016neural,santy-etal-2019-inmt}. IMT systems utilize MT systems to complete the rest of a translation after human translators editing a prefix translation \cite{alabau2014casmacat,zhao+:2020:balancing}.  For most IMT systems, the core to achieve this completion is prefix-constrained decoding \cite{wuebker2016models}. 

Another sentence-level autocompletion method, lexically constrained decoding (LCD) \cite{hokamp2017lexically, post2018fast}, recently attracts lots of attention \cite{hasler2018neural,susanto2020lexically,kajiwara2019negative}. Compared with IMT, LCD relaxes the constraints provided by human translators from prefixes to general forms: LCD completes a translation based on some unordered words (i.e., lexical constraints), which are not necessary to be continuous \cite{hokamp2017lexically, hu-etal-2019-improved, dinu2019training, song-etal-2019-code}. Although it does not need additional training, its inference is typically less efficient compared with the standard NMT. Therefore, other works propose efficient methods~\cite{li+:2020:neural,song-etal-2019-code} by using lexical constraints in a soft manner rather than a hard manner as in LCD. 

\paragraph{Word-level Autocompletion} Word-level autocompletion for CAT is less studied than sentence-level autocompletion.
\citet{langlais2000transtype, santy-etal-2019-inmt} consider to complete a target word based on human typed characters and a translation prefix. But they require the target word to be the next word of the translation prefix, which limits its application. In contrast, in our work the proposed word-level autocompletion is more general and can be applied to real-world scenarios such as post-editing~\citep{vasconcellos1985spanam,green2013efficacy} and LCD, where human translators need to input some words (corrections or constraints). 
\citet{huang2015new} propose a method to predict a target word based on human typed characters, however, this method only uses the source side information and does not consider translation context, leading to limited performance compared with our work.

\paragraph{Others} Our work may also be related to previous works in input method editors (IME) \cite{huang2018moon, lee2007language}. However, they are in the monolingual setting and not capable of using the useful multilingual information.




\section{Task and Benchmark}


In this section, we first describe why we need word-level autocompletion in real-world CAT scenarios. We then present the details of the GWLAN task and the construction of benchmark.

\paragraph{Why GWLAN?} 

Word level autocompletion is beneficial for improving input efficiency~\cite{langlais2000transtype}. 
Previous works assume that the translation context should be a prefix and the desired word is next to the prefix as shown in Figure \ref{fig:illustration} (b), where the context is ``We asked two" and the desired word is ``specialists". 
However, in some real-world CAT scenarios such as post-editing and lexically constrained decoding, translation context may be discontinuous and the input words (corrections or lexical constraints) are not necessarily conjunct to the translation context. As shown in Figure \ref{fig:illustration} (c), the context is ``We $\cdots$ their opinion" and the human typed characters ``sp" is conjunct to neither ``We'' nor ``their'' in the context. Therefore, existing methods can not perform well on such a general scenario. This motivates us to propose a general word-level autocompletion task for CAT.




\subsection{Task Definition} 
Suppose $\seq{x}=(x_1, x_2, \dots, x_m)$ is a source sequence, $\seq{s}=(s_1, s_2, \dots, s_k)$ is a sequence of human typed characters, and a translation context is denoted by $\seq{c}=(\seq{c}_l, \seq{c}_r)$, where $\seq{c}_l=(c_{l,1}, c_{l,2}, \dots, c_{l,i})$, $\seq{c}_r=(c_{r,1}, c_{r,2}, \dots, c_{r,j})$. The translation pieces $\seq{c}_l$ and $\seq{c}_r$ are on the left and right hand side of $\seq{s}$, respectively. Formally, given a source sequence $\seq{x}$, typed character sequence $\seq{s}$ and a context $\seq{c}$, the \textit{general word-level autocompletion} (GWLAN) task aims to predict a target word $w$ which is to be placed in the middle between $\seq{c}_l$ and $\seq{c}_r$ to constitute a partial translation. Note that in the partial translation consisting of $\seq{c}_l$, $w$ and $\seq{c}_r$, $w$ is not necessary to be consecutive to $c_{l,i}$ or $c_{r,1}$. 
For example, in Figure 1 (c), 
$\seq{c}_l=(\text{``We"},)$,  $\seq{c}_r=(\text{``their"},\text{``option"},\text{``."})$, $\seq{s}=(\text{``sp"},)$, the GWLAN task is expected to predict $w=\text{``specialists"}$ to constitute a partial translation ``We $\cdots$ specialists  $\cdots$ their opinion.'', where ``$\cdots$" represents zero, one, or more words (i.e., the two words before and after it are not necessarily consecutive).

To make our task more general in real-world scenarios, we assume that the left context $\seq{c}_{l}$ and right context $\seq{c}_{r}$ can be empty, which leads to the following four types of context:
\begin{itemize}[wide=0\parindent,noitemsep, topsep=0pt]
  \item Zero-context: both $\seq{c}_{l}$ and $\seq{c}_{r}$ are empty;
  \item Suffix: $\seq{c}_{l}$ is empty;
  \item Prefix: $\seq{c}_{r}$ is empty;
  \item Bi-context: neither $\seq{c}_{l}$ nor $\seq{c}_{r}$ is empty.
\end{itemize}
With the tuple $(\seq{x}, \seq{s}, \seq{c})$, the GWLAN task is to predict the human desired word $w$.

\paragraph{Relation to most similar tasks} Some similar techniques have been explored in CAT. \citet{green2014human} and \citet{knowles2016neural} studied a autocompletion scenario called translation prediction (TP), which provides suggestions of the next word (or phrase) given a prefix. Besides the strict assumption of translation context (i.e., prefix here), compared with GWLAN, another difference is that the information of human typed characters is ignored in their setting. There also exist some works that consider the human typed sequences \cite{huang2015new, santy-etal-2019-inmt}, but they only consider a specific type of translation contexts. \citet{huang2015new} propose to complete a target word based on the zero-context assumption. Despite its flexibility, this method is unable to explore translation contexts to improve the autocompletion performance. The word-level autocompletion methods in \citet{langlais2000transtype, santy-etal-2019-inmt} have the same assumption as TP, which impedes the use of their methods under the scenarios like post editing and lexically constrained decoding, where human inputs are not necessarily conjunct to the variety of translation contexts.

\subsection{Benchmark Construction\label{sec:benchmark}}

To set up a benchmark, firstly we should create a large scale dataset including tuples of $(\seq{x}, \seq{s}, \seq{c}, w )$ for training and evaluating GWLAN models. Ideally, we may hire professional translators to manually annotate such a dataset, but it is too costly in practice. 
Therefore, in this work, we propose to automatically construct the dataset from parallel datasets which is originally used in automatic machine translation tasks. The procedure for constructing our data is the same for train, validation, and test sets. And we construct a dataset for each type of translation context.

Assume we are given a parallel dataset $ \{(\seq{x}^i, \seq{y}^i)\}$, where $\seq{y}^i$ is the reference translation of $\seq{x}^i$. 
Then, we can automatically construct the data $\seq{c}^i$ and $\seq{s}^i$ by randomly sampling from $\seq{y}^i$. We first sample a word $w=\seq{y}^i_k$ and then demonstrate how to extract $\seq{c}^i$ for different translation contexts:

\begin{itemize}[wide=0\parindent,noitemsep, topsep=0pt]
    \item Zero-context: both $\seq{c}_l$ and $\seq{c}_r$ are empty;
    \item Suffix: randomly sample a translation piece $\seq{c}_r=\seq{y}_{p_{r, 1}:p_{r, 2}}$ from  $\seq{y}$, where $k < p_{r, 1} < p_{r, 2}$. The $\seq{c}_l$ is empty here; 
    \item Prefix: randomly sample a translation piece $\seq{c}_l=\seq{y}_{p_{l, 1}:p_{l, 2}}$ from  $\seq{y}$, where $p_{l, 1} < p_{l, 2} < k$. The $\seq{c}_r$ is empty here; 
    \item Bi-context: sample $\seq{c}_l$ as in prefix, and sample $\seq{c}_r$ as in suffix.
\end{itemize}

Then we have to simulate the human typed characters $\seq{s}$ based on $w$. For languages like English and German, we sample a position $p$ from the character sequence and the human input $\seq{s} = w_{1:p}$, where $1\leq p < L_w$. For languages like Chinese, the human input is the phonetic symbols of the word, since the word cannot be directly typed into the computer. Therefore, we have to convert $w$ to phonetic symbols that are characters in alphabet and sample $\seq{s}$ from phonetic symbols like we did on English.


\paragraph{Evaluation Metric}
To evaluate the performance of the well-trained models, we choose accuracy as the evaluation metric:
\begin{equation}
    \mathrm{Acc} = \frac{N_{match}}{N_{all}},
\end{equation}
where $N_{match}$ is the number of words that are correctly predicted  and $N_{all}$ is the number of testing examples.

\section{Proposed Approach}

\begin{figure}
  \begin{center}
      \includegraphics[width=6. cm]{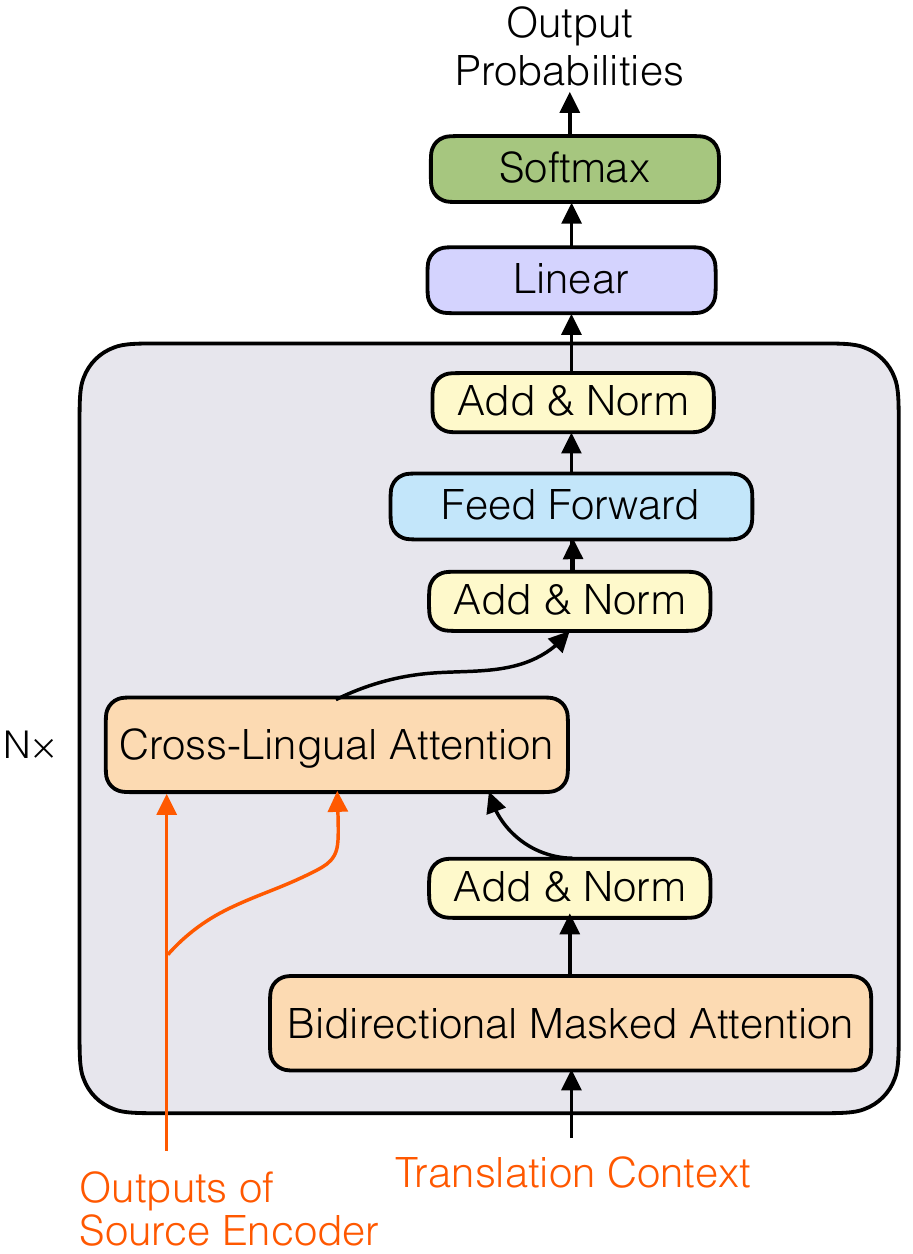}
      \caption{Cross-lingual encoder of the WPM.\label{fig:decoder}}
  \end{center}
\end{figure}

Given a tuple $(\seq{x}, \seq{c}, \seq{s})$, our approach decomposes the whole word autocompletion process into two parts: model the distribution of the target word $w$ based on the source sequence $\seq{x}$ and the translation context $\seq{c}$, and find the most possible word $w$ based on the distribution and human typed sequence $\seq{s}$. Therefore, in the following subsections, we firstly propose a word prediction model (WPM) to define the distribution $p(w|\seq{x}, \seq{c})$ of the target word $w$ (\S 4.1). Then we can treat the human input sequence $\seq{s}$ as soft constraints or hard constraints to complete $\seq{s}$ and obtain the target word $w$ (\S 4.2). Finally, we present two strategies for training and inference (\S 4.3).
\subsection{Word Prediction Model}

\begin{figure*}[t]
  \begin{center}
      \includegraphics[width=1.5\columnwidth]{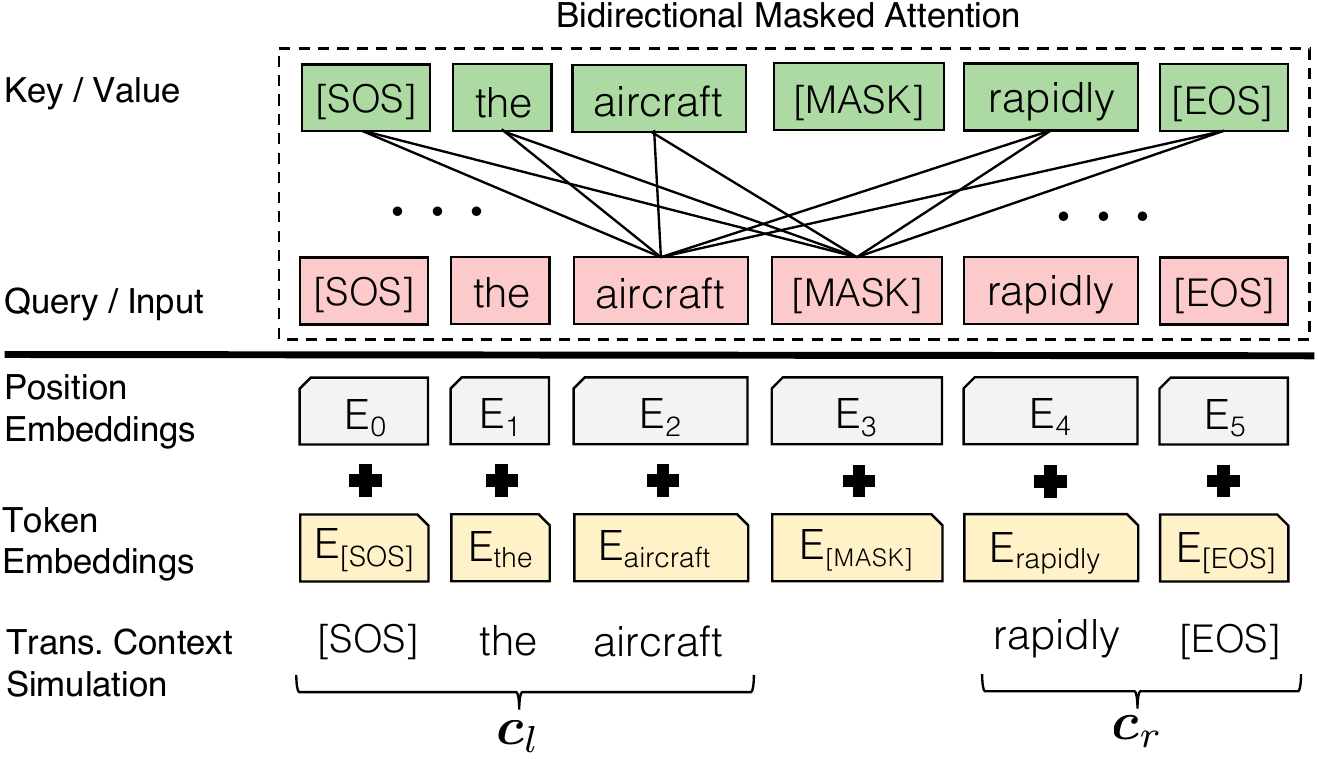}
      \caption{The input representation of our model and architecture of Bidirectional Masked Attention. The input embeddings are the sum of the token embeddings and position embeddings. \texttt{[MASK]} represents the potenial target word in this translation context. \label{fig:attention}}
  \end{center}
\end{figure*}

The purpose of WPM is to model the distribution $p(w|\seq{x}, \seq{c})$. More concretely, we will use a single placeholder \texttt{[MASK]} to represent the unknown target word $w$, and use the representation of \texttt{[MASK]} learned from WPM to predict it. Formally, given the source sequence $\seq{x}$, and the translation context $\seq{c}=(\seq{c}_l$, $\seq{c}_r)$, the possibility of the target word $w$ is:
\begin{equation}\label{eq:objective}
 P(w|\seq{x}, \seq{c}_l, \seq{c}_r;\theta) = \mathrm{softmax} \left(\phi (h) \right)[w]
\end{equation}
where $h$ is the representation of \texttt{[MASK]}, $\phi$ is a linear network that projects the hidden representation $h$ to a vector with dimension of target vocabulary size $V$, and $\mathrm{softmax}(d)[w]$ takes the component regarding to $w$ after the softmax operation over a vector $d \in \mathbb{R}^{V}$. 

Inspired by the attention-based architectures \cite{vaswani2017attention, devlin2018bert}\footnote{Because the use of attention-based models has become ubiquitous recently, we omit an exhaustive background description of the model and refer readers to \citet{vaswani2017attention} and \citet{devlin2018bert}.}, we use a dual-encoder architecture to learn the representation $h$ based on source sequence $\seq{x}$ and translation context $\seq{c}$. Our model has a source encoder and a cross-lingual encoder.  The source encoder of WPM is the same as the Transformer encoder, which is used to encode the source sequence $\seq{x}$. As shown in Figure \ref{fig:decoder}, the output of source encoder will be passed to the cross-lingual encoder later. The cross-lingual encoder is similar to the Transformer decoder, while the only difference is that we replace the auto-regressive attention (ARA) layer by a bidirectional masked attention (BMA) module, due to that the ARA layer cannot use the leftward information flow (i.e., $\seq{c}_r$).  

Specifically, the BMA module is built by a multiple-layer self attention network. As shown in Figure \ref{fig:attention}, in each layer of BMA, each token in the attention query can attend to all words in translation context $\seq{c}_l$ and $\seq{c}_r$. 
In addition, the input consists of three parts, the \texttt{[MASK]} token, and translation contexts $\seq{c}_l$ and $\seq{c}_r$, as illustrated in Figure \ref{fig:attention}.  Note that its position embeddings \texttt{E} are only used to represent the relative position relationship between tokens. Taking the sentence in Figure \ref{fig:attention} as an example, \texttt{E}$_3$ does not precisely specify the position of the target word $w$ but roughly indicates that $w$ is on the right-hand-side of $\seq{c}_l$ and on the left-hand-side of $\seq{c}_r$. Finally, the representation of \texttt{[MASK]} as learnt by BMA will be passed to \texttt{Add \& Norm} layer as shown in Figure \ref{fig:decoder}.

\subsection{Human Input Autocompletion}
After learning the representation $h$ of the \texttt{[MASK]} token, there are two ways to use the human input sequence $\seq{s}$ to determinate the human desired word. Firstly, we can learn the representation of $\seq{s}$ and use it as a soft constraint while predicting word $w$. Taking the sentence in Figure \ref{fig:attention} as an example, supposing the human typed sequence is $\seq{s} = \text{``\texttt{des}"}$, we can use an RNN network to learn the representation of \texttt{des} and concatenate it with $h$ to predict the word \texttt{descending}. Alternatively, we can use \texttt{des} as a hard constraint:
$$
P_{\seq{s}}[w] = \begin{cases}
\frac{P(w \mid \seq{x}, \seq{c}; \theta)}{Z} ,& \text{if } w \text{ starts with } \seq{s} \\
0,              & \text{otherwise.}
\end{cases}
$$ where $P(\cdot|\cdot)$ is the probability distribution defined in Eq.~\eqref{eq:objective} and  $Z$ is the normalization term independent on $w$. 
Then we pick $w^* = \argmax_{w} P_{\seq{s}}[w]$ as the most possible word. In our preliminary experiments, the performances of these two methods are comparable, and there is no significant gain when we use them together. One main reason is that the model can already learn the starts-with action precisely in the soft constraint method. Therefore, we propose to use the human inputs as hard constraints in our later experiments, because of the method's efficiency and simplicity.

\subsection{Training and Inference Strategy\label{sec:train}}
Suppose $\mathcal{D}$ denotes the training data for GWLAN, i.e., a set of tuples $(\seq{x}, \seq{c}, \seq{s}, w)$. 
Since there are four different types of context in $\mathcal{D}$ as presented in \S 3, we can split $\mathcal{D}$ into four subsets $\mathcal{D}_{\text{zero}}$, $\mathcal{D}_{\text{prefix}}$, $\mathcal{D}_{\text{suffix}}$ and $\mathcal{D}_{\text{bi}}$. 
To yield good performances on those four types of translation context, we also propose two training strategies. The inference strategy differs accordingly. 

\paragraph{Strategy 1: One Context Type One Model} For this strategy, we will train a model for each translation context, respectively. Specifically, for each type of context $t\in \{\text{zero}, \text{prefix}, \text{suffix}, \text{bi}\}$, we independently train one model $\theta_t$ by minimizing the following loss $\mathcal{L}(\mathcal{D}_t, \theta)$:
\begin{equation}
    \mathcal{L}(\mathcal{D}_t; \theta) = \frac{1}{|\mathcal{D}_t|} \sum_{(\seq{x}, \seq{c}, \seq{s}, w) \in \mathcal{D}_t}\log P(w|\seq{x}, \seq{c}; \theta),
    \label{eq:loss}
\end{equation}
where $P(w|\seq{x}, \seq{c}; \theta)$ is the WPM model defined in Eq.~\ref{eq:objective}, $|\mathcal{D}_t|$ is the size of training dataset $\mathcal{D}_t$, and $t$ can be any type of translation context. 
In this way, we actually obtain four models in total after training. In the inference process, for each testing instance $(\seq{x}, \seq{c}_l, \seq{c}_r, \seq{s})$, we decide its context type $t$ in terms of $\seq{c}_l$ and $\seq{c}_r$ and then use $\hat{\theta}_t$ to predict the  word $w$.

\paragraph{Strategy 2: Joint Model} The separate training strategy is straightforward. However, it may also make the models struck in the local optimal. To address these issues, we also propose a joint training strategy, which has the ability to stretch the model out of the local optimal once the parameters is over-fitting on one particular translation context. 
Therefore, using the joint training strategy, we train a single model for all types of translation context by minimizing the following objective:
\begin{multline*}
        \mathcal{L}(\mathcal{D}; \hat{\theta}) = \mathcal{L}(\mathcal{D}_{\text{zero}}; \hat{\theta}) + \mathcal{L}(\mathcal{D}_{\text{prefix}}; \hat{\theta}) + \\  \mathcal{L}(\mathcal{D}_{\text{suffix}}; \hat{\theta}) + \mathcal{L}(\mathcal{D}_{\text{bi}}; \hat{\theta})
\end{multline*}
\noindent where each $\mathcal{L}(\mathcal{D}_{t}; \theta)$ is as defined in Eq.~\ref{eq:loss}. 
In this way, we actually obtain a single model $\hat{\theta}$ after training. In the inference process, for each testing instance $(\seq{x}, \seq{c}_l, \seq{c}_r, \seq{s})$ we always use $\hat{\theta}$ to predict the target word $w$.

\section{Experiments}

\begin{table*}[t]
\resizebox{2.08\columnwidth}{!}{
\begin{tabular}{@{}l|l||cc||cc||cc||cc@{}}
\toprule
  \multirow{2}{*}{\#}   &     \multirow{2}{*}{Systems}     & \multicolumn{2}{c||}{Zh$\Rightarrow$En}                                & \multicolumn{2}{c||}{En$\Rightarrow$Zh}                                & \multicolumn{2}{c||}{De$\Rightarrow$En} & \multicolumn{2}{c}{En$\Rightarrow$De} \\ \cline{3-10}
    &       & \multicolumn{1}{c}{NIST05} & \multicolumn{1}{c||}{NIST06} & \multicolumn{1}{c}{NIST05} & \multicolumn{1}{c||}{NIST06} & NT13        & NT14       & NT13        & NT14       \\\hline\hline
1 & \textsc{TransTable} & 41.40                      & 39.78                      & 28.00                      & 26.99                      &     37.43        &       36.64     &       32.99      &    31.12        \\
2 & \textsc{Trans-PE}   & 34.51                      & 35.50                      & 32.23                      & 34.88                      &        34.45     &      33.02      &      31.51      &          30.65  \\
3 & \textsc{Trans-NPE}  & 35.97                      & 36.78                      & 34.31                      & 36.19                      &         36.69    &      36.01      &       33.25      &        31.30    \\
4 &  \textsc{WPM-Sep}    & 54.15                      & 55.04                      & 53.30                      & 53.67                      &        56.93     &     55.67       &      54.54       &    51.46        \\
5 & \textsc{WPM-Joint} & \textbf{55.54}                      & \textbf{55.85}                      & \textbf{53.64}                      & \textbf{54.25}                             &       \textbf{57.84}     &    \textbf{56.75}   &    \textbf{56.91}     &      \textbf{52.68}      \\ \bottomrule
\end{tabular}
} \caption{\label{tab:all} The main results of different systems on Chinese-English and German-English datasets. The results in this table are the averaged accuracy on four translation contexts (i.e., prefix, suffix, zero-context, and bi-context).}
\end{table*}

\begin{table*}[t]
  \resizebox{2.08\columnwidth}{!}{
\begin{tabular}{@{}l|l||ccccc||ccccc@{}}
\toprule
  \multirow{2}{*}{\#} &    \multirow{2}{*}{Systems}         & \multicolumn{5}{c||}{Zh$\Rightarrow$En}                                                                                                                              & \multicolumn{5}{c}{En$\Rightarrow$Zh} \\ \cline{3-12}
 &            & \multicolumn{1}{l}{Prefix} & \multicolumn{1}{c}{Suffix} & \multicolumn{1}{c}{Zero} & \multicolumn{1}{c}{Bi} & \multicolumn{1}{c||}{Avg.} & \multicolumn{1}{c}{Prefix} & \multicolumn{1}{c}{Suffix} & \multicolumn{1}{c}{Zero} & \multicolumn{1}{c}{Bi} & \multicolumn{1}{c}{Avg.} \\\hline\hline
1  & \textsc{TransTable} & 41.91                      & 44.99                      & 44.19                            & 43.28                          & 43.59                   & 29.73                      & 32.80                       & 29.73                            & 29.61                          & 30.46                   \\
2  & \textsc{Trans-PE}   & 29.84                      & 38.61                      & 26.08                            & 48.06                          & 35.64                   & 30.64                      & 34.97                      & 22.67                            & 38.95                          & 31.80                   \\
3  & \textsc{Trans-NPE}  & 37.36                      & 40.43                      & 29.50                             & 44.42                          & 37.92                   & 36.10                       & 43.05                      & 32.00                               & 45.79                          & 39.23                   \\
    4  & \textsc{WPM-Sep}    & 58.43 & 60.59 & 53.99 & \textbf{64.46} & 59.36 & 60.02 & 61.05 & 53.76 & \textbf{64.46} & 59.82 \\
    5  & \textsc{WPM-Joint}  & \textbf{59.91} & \textbf{60.71} & \textbf{55.35} & 62.30 & \textbf{59.56} & \textbf{61.39} & \textbf{61.73} & \textbf{53.87} & 63.78 & \textbf{60.19}        \\ \bottomrule
\end{tabular}
}
\caption{\label{tab:nist02} The results of different systems on NIST02. We evaluate the performances of those systems on both Zh$\Rightarrow$En and En$\Rightarrow$Zh tasks by accuracy. }
\end{table*}

\subsection{Datasets} We carry out experiments on four GWLAN tasks including bidirectional Chinese–English tasks and German–English tasks. The benchmarks for our experiments are based on the public translation datasets. The training set for two directional Chinese–English tasks consists of 1.25M bilingual sentence pairs from LDC corpora. The toolkit we used to convert Chinese word $w$ to phonetic symbols is pypinyin\footnote{\url{https://github.com/mozillazg/python-pinyin}}. As discussed in  (\S \ref{sec:benchmark}), the training data for GWLAN is extracted from 1.25M sentence pairs. The validation data for GWLAN is extracted from NIST02 and the test datasets for GWLAN are constructed from NIST05 and NIST06. For two directional German–English tasks, we use the WMT14 dataset preprocessed by Stanford\footnote{\url{https://nlp.stanford.edu/projects/nmt/}}. The validation and test sets for our tasks are based on newstest13 and newstest14 respectively. For each dataset, the models are tuned and selected based on the validation set.

The main strategies we used to prepare our benchmarks are shown in \S\ref{sec:benchmark}. However, lots of trivial instances may be included if we directly use the uniform distribution for sampling, e.g., predicting word ``the'' given ``th''. Therefore, we apply some intuitive rules to reduce the  probability of trivial instances. For example, we assign higher probability for words with more than 4 characters in English and 2 characters in Chinese, and we require that the lengths of input character sequence $\seq{s}$ and translation contexts $\seq{c}$ should not be too long.

\subsection{Systems for Comparison} In the experiments, we evaluate and compare the performance of our methods (WPM-Sep and WPM-Joint) and a few baselines. They are illustrated below,

\paragraph{\textsc{WPM-Sep}} is our approach with the ``one context one model'' training and inference strategy in Section \S \ref{sec:train}. In other words, we train our model for each translation context separately. 
\paragraph{\textsc{WPM-Joint}} is our approach with the ``joint model'' strategy in Section \S \ref{sec:train}.
\paragraph{\textsc{TransTable}:} We train an alignment model\footnote{\url{https://github.com/clab/fast_align}} on the training set and build a word-level translation table. While testing, we can find the translations of all source words based on this table, and select out valid translations based on the human input. The word with highest frequency among all candidates is regarded as the prediction. This baseline is inspired by ~\citet{huang2015new}. 
\paragraph{ \textsc{Trans-PE}:} We train a vanilla NMT model using the Transformer-base model. During the inference process, we use the context on the left hand side of human input as the model input, and return the most possible words based on the probability of valid words selected out by the human input. This baseline is inspired by ~\citet{langlais2000transtype, santy-etal-2019-inmt}.
\paragraph{\textsc{Trans-NPE}:} As another baseline, we also train an NMT model based on Transformer, but without position encoding on the target side. While testing, we use the averaged hidden vectors of all the target words outputted by the last decoder layer to predict the potential candidates.


\subsection{Main Results}

Table \ref{tab:all} shows the main results of our methods and three baselines on the test sets of Chinese-English and German-English datasets. It is clear from the results that our methods \textsc{WPM-Sep} and \textsc{WPM-Joint} significantly outperform the three baseline methods. Results on Row 4 and Row 5 of Table \ref{tab:all} also show that the \textsc{WPM-Joint} method, which uses a joint training strategy to optimize a single model, achieves better overall performance than \textsc{WPM-Sep}, which trains four models for different translation contexts respectively. In-depth analysis about the two training strategies is presented in the next section.

The method \textsc{Trans-PE}, which assumes the human input is the next word of the given context, behaves poorly under the more general setting. As the results of \textsc{Trans-NPE} show, when we use the same model as \textsc{Trans-PE} and relax the constraint of position by removing the position encoding, the accuracy of the model improves. One interesting finding is that the \textsc{TransTable} method, which is only capable of leveraging the zero-context, achieves good results on the Chinese-English task when the target language is English. However, when the target language is Chinese, the performance of \textsc{TransTable} drops significantly.

\section{Experimental Analysis}
\subsection{Effects on Different Translation Context}

In this section, we presents more detailed results on the four translation contexts and analyze the features of GWLAN. These analyses can help us to better understand the task and propose effective approaches in the future.

\paragraph{Separate Training VS. Joint Training} 
Compared with \textsc{WPM-Sep}, \textsc{WPM-Joint} shows two advantages. On one hand,  even there is only one model, \textsc{WPM-Joint} yields better performances than \textsc{WPM-Sep}, enabling simpler deployment. This may be caused by that training on multiple related tasks can force the model learn more expressive representations, avoiding over-fitting. On the other hand, the variance of results on different translation contexts of \textsc{WPM-Joint} is smaller, which can provide an more steady autocompletion service. From the viewpoint of joint training, the lower variance may be caused by that \textsc{WPM-Joint} spends more efforts to minimize the one with maximal risk (i.e., zero-context), although sometimes it may slightly sacrifice the task with minimal risk (i.e., bi-context).

The results of \textsc{WPM-Sep} and \textsc{WPM-Joint} also have some shared patterns. Firstly, the performances of the two methods on prefix and suffix translation contexts are nearly the same. Although the prefix and suffix may play different roles in the SVO language structure, they have little impact on the the autocompletion accuracy using our method. Moreover, among the results on four translation contexts, the performances on bi-context are better than prefix and suffix, and prefix and suffix are better than zero-context. This finding shows that more context information can help to reduce the uncertainty of human desired words.

\paragraph{Comparison with baselines}

The \textsc{Trans-PE} method in previous works is more sensitive to the position of human input. The statistical results shows that the averaged distances in the original sentence between the prediction words and translation contexts are various for different translation contexts, which are $7.4$, $6.5$, $14.1$, and $3.2$ for prefix, suffix, zero-context, and bi-context, respectively. When the desired words are much closer to the context, \textsc{Trans-PE} can achieve better performances. Moreover, \textsc{Trans-PE} can achieve more than 80 accuracy scores when the prediction word is the next word of the given prefix, however, its performance drops significantly when the word is not necessarily conjunct to the prefix. We can also find that \textsc{Trans-NPE}, which removes the position information of target words, achieves better overall performances compared with \textsc{Trans-PE}.

In contrast, the performance of \textsc{TransTable} is less affected by the position of the prediction words, which is demonstrated by the low variances on both tasks in Table \ref{tab:nist02}. The results of \textsc{TransTable} have also surprised us, which achieves more than 41 accuracy scores on the Zh$\Rightarrow$En task. This observation shows the importance of alignment and the potential of statistical models. Compared with the results on the Zh$\Rightarrow$En task, the overall accuracy on En$\Rightarrow$Zh task is much lower, likely due to that the number of valid words after filtered by the human input on Chinese is much more than that on English.
Therefore, it is easier for \textsc{TransTable} to determine the human desired words in English.

\subsection{Robustness on Noisy Contexts} In this work, the translation contexts are simulated using the references. However, in real-world scenarios, translation contexts may not be perfect, i.e., some words in the translation contexts may be incorrect. In this section, we evaluate the robustness of our model on noisy contexts. We first use the translation table constructed by \textsc{TransTable} to find some target words that share the same source words with the original target words, and then use those found words as noise tokens.

\begin{figure}
\begin{center}
   \includegraphics[width=1.0\linewidth]{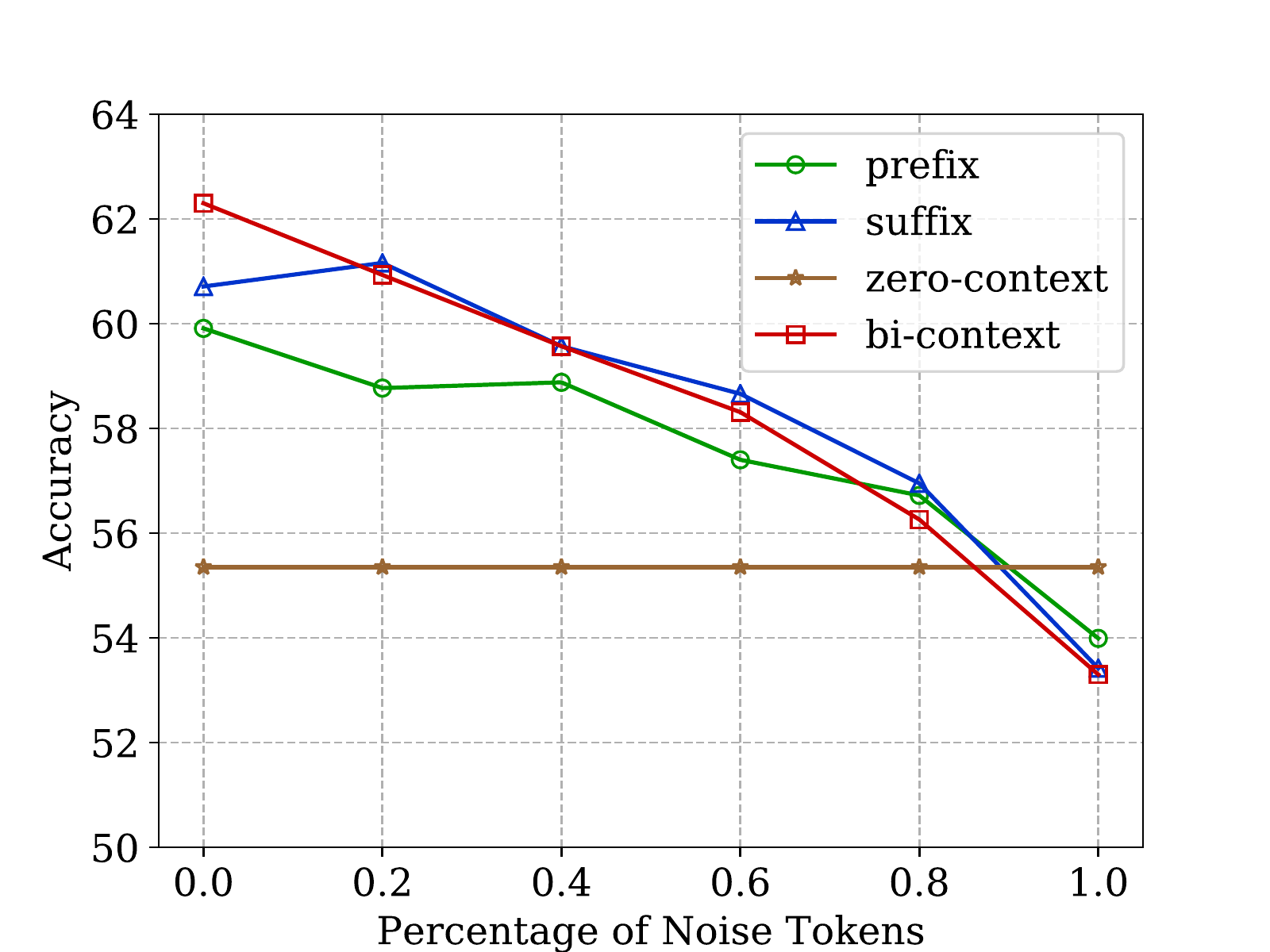}
\end{center}
   \caption{Robustness Analysis. The x-axis represents the percentage of words  that have been replaced by noise tokens in NIST02. The model used for this analysis is the \textsc{WPM-Joint}, which is trained on the Zh$\Rightarrow$En task without noisy translation context. \label{tab:noise}}
\end{figure}

The robustness results are shown in Figure \ref{tab:noise}. For all the translation context types except for zero-context, the performance drops slowly when the percentage of noise tokens increases. However, even with 80\% words in the context, the performance of \textsc{WPM-Joint} outperforms the case of zero-context, which shows that our \textsc{WPM-Joint} method is noise tolerant.

\subsection{Discussion}

In this work, we formalize the task as a classification problem. However, the generation formalization also deserves to be explored in the future. For example, the generation may happen in two circumstances: word-level completion based on subwords, and phrase-level completion. In the first case, although the autocompletion service provided for human translators is word-level, in the internal system we can generate a sequence of subwords \cite{sennrich2015neural} that satisfy the human typed characters, and provide human translators with the merged subwords. This subword sequence generation can significantly alleviate the OOV issue in the word-level autocompletion. In the phrase-level autocompletion case, if we can predict more than one desired words, the translation efficiency and experience may be improved further.
We would like to leave it as future work.

It is also worth noting that we did not conduct human studies in this work. We think evidences in previous work can already prove the effectiveness of word-level autocompletion when assisting human translators. For example, TransType \cite{langlais2000transtype} is a simple rule-based tool that only considers the prefix context, but the majority of translators said that TransType improved their typing speed a lot. \citet{huang2015new} hired 12 professional translators and systematically evaluate their word autocompletion tool based on zero-context. Experiments show that the more keystrokes are reduced, the more time can be saved for translators. Since the prediction accuracy is highly correlated with the keystrokes, we think higher accuracy will make translators more productive. That is the main reason that we use accuracy to automatically evaluate the model performance. Besides, the automatic evaluation metric also makes the GWLAN task easier to follow.

\section{Conclusion}

We propose a General {W}ord-{L}evel {A}utocompletio{N} (GWLAN) task for computer-aided translation (CAT). In our setting, we relax the strict constraints on the translation contexts in previous work, and abstract four most general translation contexts used in real-world CAT scenarios.
We propose two approaches to address the variety of context types and weak position information issues in GWLAN.
To support automatic evaluation and to ensure a convenient and fair comparison among different methods, we construct a benchmark for the task.
Experiments on this benchmark show that our method outperforms baseline methods by a large margin on four datasets. We believe that this benchmark to be released will push forward future research in CAT.

\section*{Acknowledgments}
We would like to thank three anonymous reviewers for their invaluable discussions on this work. The corresponding is Lemao Liu. 

\bibliographystyle{acl_natbib}
\bibliography{acl2021}


\end{document}